\documentclass[conference]{IEEEtran}
\IEEEoverridecommandlockouts
\usepackage{cite}
\usepackage{amsmath,amssymb,amsfonts}
\usepackage{algorithmic}
\usepackage{graphicx}
\usepackage{textcomp}
\usepackage{xcolor}
\usepackage{booktabs}
\usepackage{dblfloatfix}
\def\BibTeX{{\rm B\kern-.05em{\sc i\kern-.025em b}\kern-.08em
    T\kern-.1667em\lower.7ex\hbox{E}\kern-.125emX}}
\begin{document}

\title{Exploring Lip Segmentation Techniques in Computer Vision: A Comparative Analysis\\
}

\author{\IEEEauthorblockN{Pietro B. S.  Masur}
\IEEEauthorblockA{\textit{Neurobots} \\
Recife, Brazil \\
pietro.masur@neurobots.com.br}
\and
\IEEEauthorblockN{Francisco Braulio Oliveira}
\IEEEauthorblockA{\textit{CESAR} \\
Recife, Brazil \\
fbso@cesar.org.br}
\and
\IEEEauthorblockN{Lucas Moreira Medino}
\IEEEauthorblockA{\textit{Neurobots} \\
Recife, Brazil \\
lucas.medino@neurobots.com.br}
\and
\IEEEauthorblockN{Emanuel Huber}
\IEEEauthorblockA{\textit{CESAR} \\
Recife, Brazil \\
ehs@cesar.org.br}
\and
\IEEEauthorblockN{Milene Haraguchi Padilha}
\IEEEauthorblockA{\textit{Grupo Boticário} \\
Curitiba, Brazil \\
milenehp@grupoboticario.com.br}
\and
\IEEEauthorblockN{Cassio de Alcantara}
\IEEEauthorblockA{\textit{Grupo Boticário} \\
Curitiba, Brazil \\
cassio.alcantara@grupoboticario.com.br}
\and
\IEEEauthorblockN{Renata Sellaro}
\IEEEauthorblockA{\textit{CESAR} \\
Recife, Brazil \\
rfs@cesar.org.br}
}

% \author{
%   Pietro Masur$^*$,
%   Francisco Braulio Oliveira$^\dagger$,
%   Cassio de Alcantara$^\ddagger$,
%   Lucas Medino$^*$, \\
%   Emanuel Huber$^\dagger$,
%   Milene Haraguchi Padilha$^\ddagger$,
%   Renata Sellaro$^\dagger$ \\
%   \small $^*$Neurobots, Recife, Pernambuco, Brazil \\
%   \small $^\dagger$Centro de Estudos e Sistemas Avançados do Recife (CESAR), Recife, Pernambuco, Brazil \\
%   \small $^\ddagger$Grupo Boticario, Curitiba, Parana, Brazil \\
%   \small emails: \{pietro.masur, lucas.medino\}@neurobots.com.br, \{fbso, ehs\}@cesar.org.br, \{cassio.alcantara, milenehp\}@grupoboticario.com.br
% }
\IEEEoverridecommandlockouts
\IEEEpubid{\makebox[\columnwidth]{979-8-3503-4807-1/23/\$31.00~\copyright2023 IEEE\hfill} \hspace{\columnsep}\makebox[\columnwidth]{ }}

\maketitle

\IEEEpubidadjcol

\begin{abstract}

Lip segmentation is crucial in computer vision, especially for lip reading.
Despite extensive face segmentation research, lip segmentation has received limited attention.
The aim of this study is to compare state-of-the-art lip segmentation models using a standardized setting and a publicly available dataset.
Five techniques, namely EHANet, Mask2Former, BiSeNet V2, PIDNet, and STDC1, are qualitatively selected based on their reported performance, inference time, code availability, recency, and popularity.
The CelebAMask-HQ dataset, comprising manually annotated face images, is used to fairly assess the lip segmentation performance of the selected models.
Inference experiments are conducted on a Raspberry Pi4 to emulate limited computational resources.
The results show that Mask2Former and EHANet have the best performances in terms of mIoU score.
BiSeNet V2 demonstrate competitive performance, while PIDNet excels in recall but has lower precision.
Most models present inference time ranging from 1000 to around 3000 milliseconds on a Raspberry Pi4, with PIDNet having the lowest mean inference time.
This study provides a comprehensive evaluation of lip segmentation models, highlighting their performance and inference times.
The findings contribute to the development of lightweight techniques and establish benchmarks for future advances in lip segmentation, especially in IoT and edge computing scenarios.

\end{abstract}


\section{Introduction}
\label{intro}

% Comentar o porquê segmentação labial é importante
% Falar sobre os avanços
% Falar sobre as lacunas

Computer vision techniques enable computers and systems to extract meaningful information from visual inputs such as images and videos.
Based on this information, these systems can take appropriate actions or make recommendations based on this information.
In contrast to humans, who benefit from a lifetime of contextual experiences that enables them to discern objects, estimate distances, and spot anomalies, computer vision systems aim to train machines to perform these tasks by using data and algorithms.
A system trained on large amounts of data is able to inspect products or monitor manufacturing processes to evaluate thousands of items or procedures per minute, detecting errors or problems faster and more accurately than humans.
Image segmentation is a fundamental task in computer vision, involving the partitioning of an image into distinct regions or the identification of objects.
One particular area of interest is lip segmentation, which has important applications in videoconferencing, lip reading, and low bit rate coding communication systems. %TODO ADICIONAR REFERENCIA
Despite the extensive research in lip segmentation \cite{key_ponts_seg} \cite{skodras2011unconstrained} \cite{le2016novel}, there is a lack of comparative analysis of publicly available machine learning models.
Existing publications focus on models trained and evaluated on different image datasets, some of which are not publicly available.
Additionally, the absence of standardized metrics further hinders the comparison of results reported in different publications.
Therefore, there is a need for a systematic evaluation and comparison of lip segmentation models.
Such an analysis would provide valuable insights into their performance across different datasets and establish benchmarks for future advances in the field.

This study presents a comparative analysis of machine learning and deep learning models for lip segmentation.
Our objective is to utilize a publicly available dataset to train and test several advanced models, thereby comparing their performance in a standardized setting.
Additionally, we aim to assess the inference time of each model on limited hardware.
This evaluation is relevant for scenarios such as edge computing and IoT applications.

The paper is structured as follows: Section \ref{related} provides an overview of related work.
Section \ref{method} outlines the methodology for model selection and provides specific comparison settings.
Section \ref{results} presents the obtained results.
Section \ref{discussion} provides a comprehensive discussion.
Finally, Section \ref{conclusion} concludes the paper and suggests future research directions.

\vspace{10pt}

\section{Related Work}
\label{related}

% Pesquisar sobre papers que fazem comparativos de modelos de segmentação (labial, se tiver algum)
% Incluir uma introdução sobre como buscamos esses modelos
% Incluir os modelos que o pietro for adicionar
% Remover os que não forem entrar

\begin{figure*}[!h]
    \centering
	\includegraphics[width=0.8\textwidth,keepaspectratio]{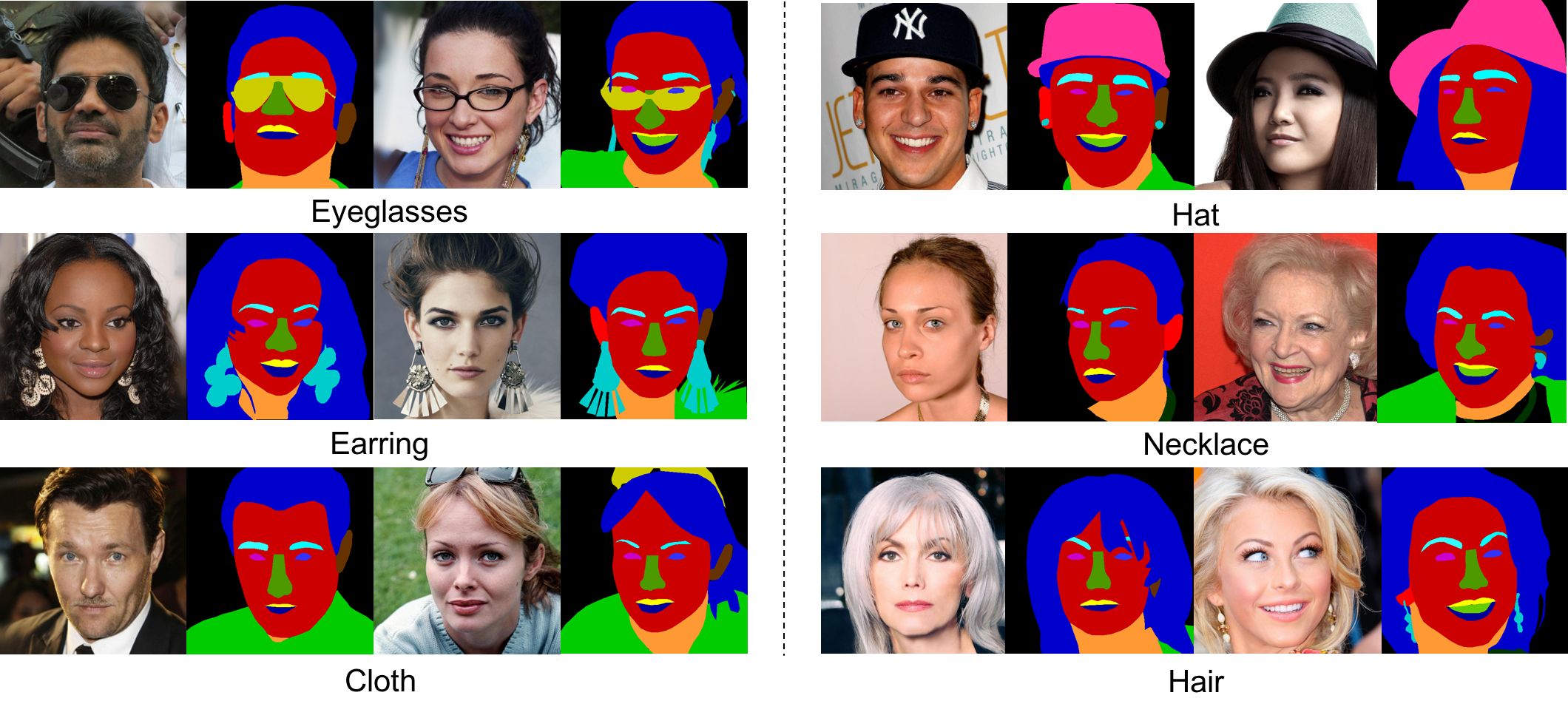}
	\caption{Sample images from the CelebAMask-HQ dataset. Several images in this dataset feature props, adding components of variability to the data.}
	\label{fig:celebamask}
\end{figure*}

Segmentation algorithms have garnered considerable attention in recent academic research, with a specific focus on facial element segmentation \cite{khan2020face}. Early studies have shown interest in lip segmentation and have explored computationally inexpensive techniques, such as Fuzzy clustering \cite{liew2003segmentation} and k-means clustering \cite{skodras2011unconstrained}, to distinguish lip-associated pigmented regions from non-lip regions. Other research efforts have concentrated on segmenting human lips based on contour and shape features. Le and Savvides introduced the Shape Constrained Feature-based Active Contour (SC-FAC) model \cite{le2016novel}, which combines feature-based active contour (FAC) and prior shape constraints (CS) to achieve precise segmentation.

With the advancement of technologies, such as deep learning, that enable the utilization of more extensive computational resources in computer vision models, the demand for large and well-labeled databases has increased. These databases are crucial for training complex models and evaluating their performance on test partitions. One widely employed database for training facial segmentation models is the CelebAMask-HQ dataset \cite{karras2018progressive}. This dataset comprises 30,000 high-resolution facial images that have been manually annotated for different facial components. The dataset follows the semantic segmentation paradigm, assigning a distinct class to each pixel. Fig. \ref{fig:celebamask} illustrates samples of images and corresponding masks from the CelebAMask-HQ dataset.

In recent years, several new models have been introduced, bringing forth new possibilities in the field of lip segmentation. Xu et al. \cite{xu2023pidnet} introduces a three-branch network architecture designed for real-time semantic segmentation tasks. While two-branch networks have proven effective, they suffer from the drawback of detailed features being overwhelmed by contextual information, limiting segmentation accuracy. Drawing a connection between Convolutional Neural Networks (CNNs) and Proportional-Integral-Derivative (PID) controllers, this study reveals that a two-branch network is equivalent to a Proportional-Integral (PI) controller, which exhibits overshoot issues. To address this problem, PIDNet is proposed, featuring three branches for parsing detailed, contextual, and boundary information, with boundary attention employed to guide the fusion of detailed and contextual branches.

\begin{table*}[h]
    \caption{Performance summary of the selected methods.}
    \label{tab:methods_summary}
    \centering
    \resizebox{0.75\textwidth}{!}{%
\centering
\begin{tabular}{@{}lllllllll@{}}
\toprule
\textbf{Method} & \textbf{mIoU} & \textbf{Dataset} & \textbf{FPS} & \textbf{Device} & \textbf{\begin{tabular}[c]{@{}l@{}}Crop\\ Size \end{tabular}} & \textbf{Code} & \textbf{Publication} & \textbf{Citations} \\ \midrule
EHAnet & 78.2 & CelebAMask & 55 & GTX 1080Ti & 256x256 & Yes & 2020 & 20 \\
Mask2Former & 57.7 & ADE20K & 9 & A100 & 640x640 & Yes & 2022 & 467 \\
BiSeNet V2 & 72.6 & Cityscapes & 156 & GTX 1080Ti & 1024x1024 & Yes & 2021 & 557 \\
PIDNet & 78.6 & Cityscapes & 93 & A100 & 1024x1024 & Yes & 2023 & 18 \\
STDC1 & 71.9 & Cityscapes & 250 & GTX 1080Ti & 1024x1024 & Yes & 2021 & 273 \\
\bottomrule
\end{tabular}
}
\end{table*}

In their study, Luo et al. \cite{luo2020ehanet}  proposed EHANet, an encoder-decoder architecture for face parsing. It incorporates deep convolutional layers and features four components: stage contextual attention mechanism (SCAM) for contextual attention, semantic gap compensation block (SGCB) for semantic gap compensation, a boundary-aware (BA) module for boundary-awareness, and a defined loss function. The model outperformed others on CelebAMask-HQ, achieving 78.2\% mIoU and 90.7\% F1 score. EHANet excelled in corner and edge detection tasks, making it promising for face parsing, especially on resource-limited devices.

The focus of BiSeNet V2 \cite{yu2021bisenet} is real-time semantic segmentation and high accuracy, preserving low-level details. The proposed architecture comprises the Detail Branch and the Semantic Branch. The Detail Branch captures low-level details through wide channels and shallow layers, generating high-resolution feature representation. Conversely, the Semantic Branch obtains high-level semantic context using narrow channels and deep layers. To facilitate the fusion of both feature representations and enhance their mutual connections, a Guided Aggregation Layer is introduced. Moreover, a booster training strategy is employed to enhance segmentation performance without increasing inference cost. For a 2,048×1,024 input, the proposed architecture achieves a Mean Intersection-over-Union (IoU) of 72.6\% on the Cityscapes test set, with a speed of 156 frames per second (FPS) on a single NVIDIA GeForce GTX 1080 Ti card.

Fan et al. \cite{fan2021rethinking} introduce the Short-Term Dense Concatenate (STDC1) network. They address the limitations of the existing BiSeNet architecture, which employs a two-stream network approach but suffers from time-consuming operations and inefficient backbones borrowed from pretrained tasks. The STDC1 network gradually reduces feature map dimensions and aggregates them for image representation, forming the basic module. In the decoder, a Detail Aggregation module is introduced to incorporate spatial information into low-level layers in a single-stream manner. The fusion of low-level features and deep features enables the prediction of final segmentation results. On the Cityscapes dataset, the STDC1 network achieves a mean Intersection-over-Union (mIoU) of 71.9\% on the test set, with an inference speed of 250.4 frames per second (FPS) on an NVIDIA GTX 1080Ti. When inferring on higher resolution images, a mIoU of 76.8\% is achieved with a speed of 97.0 FPS.

Gheng et al. \cite{cheng2022masked} introduce Mask2Former, a universal architecture for image segmentation tasks, including panoptic, instance, and semantic segmentation. It incorporates a Transformer decoder with masked attention, efficiently utilizing high-resolution features. Sequential feeding of multi-scale features to the decoder addresses small object handling. Further modifications include reordering self and cross-attention, learnable query features, and removing dropout for efficiency. Evaluation on COCO \cite{zhou2017scene}, ADE20K \cite{zhou2017scene}, Cityscapes \cite{cordts2016cityscapes}, and Mapillary Vistas \cite{neuhold2017mapillary} demonstrates state-of-the-art performance. Mask2Former achieves a new benchmark on ADE20K with a mIoU of 57.7, surpassing MaskFormer and specialized models for panoptic (57.8 PQ on COCO), instance (50.1 AP on COCO), and semantic segmentation.

Numerous methodologies have emerged in the realm of image segmentation, a selection of which merits particular attention. Bazarevsky et al. \cite{bazarevsky2019blazeface} developed a tool capable of achieving a frame rate of 200-1000 FPS on high-end mobile devices, yet its code remains unavailable for public use. Dong et al. \cite{dong2023head} offered the AFFormer method, which achieved 78 FPS and 78.7 mIoU on the Cityscapes dataset with a crop size of 1024x2048 using a V100 GPU; the code for this method is available, although it is a relatively new contribution. Similarly, the DDRNet developed by Hong et al. \cite{hong2021deep} achieved 85 FPS and 77.8 mIoU on the Cityscapes dataset with the same crop size and GPU. Lastly, Zhang et al. \cite{zhang2021knet} developed the K-Net, achieving 8 FPS and 54.3 mIoU on ADE20K with a crop size of 640x640, also using a V100 GPU. These highlighted methods illustrate the breadth of current developments in image segmentation.

\vspace{10pt}

\section{Methodology}
\label{method}

In this study, our primary objective was to identify computer vision methods that best satisfied certain criteria. Specifically, we targeted methods demonstrating high performance on well-established datasets with recognized metrics, low inference time as evidenced by a high frame rate per second (fps), available open-source code, recency, and popularity reflected in a high number of citations. This work is not a systematic literature review, therefore we focused on methods that closely aligned with our established criteria. 

We undertook a qualitative selection based on our best knowledge and limited it to five methods: EHAnet, Mask2Former, BiSeNet V2, PIDNet, and STDC1. The intention was to delve deeply into each method and conduct multiple experiments. These methods, discussed in the related works section, all employ semantic segmentation, where each pixel is assigned to a specific class, and everything outside the regions of interest is treated as background.

Table \ref{tab:methods_summary} presents a summary of each selected method's performance against the specified criteria. While other methods were assessed in our research, they did not match the overall quality standards of our selected techniques.

To ensure a fair comparison, we utilized the CelebAMask-HQ dataset \cite{karras2018progressive} dataset, as described in the previous section. This dataset consists of 30,000 manually annotated images depicting various parts of the human face. The dataset contemplates a variety of face poses and expressions as well as some scenarios with partial facial occlusion. For this study, we focused only on the annotations of the upper and lower lips, considering the remaining areas as background. The CelebAMask-HQ dataset was randomly divided into training (70\%, N$\approx$21,000), validation (15\%, N$\approx$4,500), and test sets (15\%, N$\approx$4,500). A total of 165 images of the CelebAMask-HQ dataset were not considered due to the lack of either upper or lower lip masks.

To train the Mask2Former, BiSeNet V2, PIDNet, and STDC1 we have used the MMsegmentation \cite{mmseg2020} library. For EHANet, we utilized the author's implementation. First, the models were trained with the hyperparameters utilized in their respective original papers. From there, we did some minor alterations to the learning rate and the training stop criteria (number of epochs). Once we had found the best setup on hyperparameters for each model, we conducted ten training rounds for each model, using different random seeds for initialization. Finally, we performed ten executions of inference on a Raspberry Pi4, simulating low computational resource availability. Table \ref{tab:model_metrics} presents the average results and standard deviations for different evaluation metrics on the test set for each of the models, as well as for the inference time. The metrics are detailed as follows:

\subsection{mean Intersection over Union (mIoU)}
The mean Intersection over Union (mIoU) is commonly used to evaluate image segmentation tasks. It measures the similarity between the predicted segmentation masks and the ground truth masks by calculating the intersection over the union for each class and then averaging the results across all classes. The formula for calculating mIoU is as follows:

\[
mIoU = \frac{1}{N} \sum_{i=1}^{N} \frac{TP_i}{TP_i + FP_i + FN_i}
\]

\text{where:}
\begin{align*}
N & \text{ represents the total number of classes} \\
TP & \text{ True positives (correctly classified foreground pixels)} \\
TN & \text{ True negatives (correctly classified background pixels)} \\
FP & \text{ False positives (background pixels incorrectly classified)} \\
FN & \text{ False negatives (foreground pixels incorrectly classified)}
\end{align*}

\subsection{Dice}

\[
\text{Dice} = \frac{2 \times |X \cap Y|}{|X| + |Y|}
\]

The Dice coefficient is a similarity measure commonly used in evaluating the performance of image segmentation algorithms. It assesses the overlap between the predicted segmentation mask and the ground truth mask and it ranges from 0 to 1, where a value of 1 indicates a perfect overlap between the two masks. $X$ represents the predicted segmentation mask and $Y$ represents the ground truth mask. The Dice coefficient ranges from 0 to 1, where a value of 1 indicates a perfect overlap between the two masks.

\subsection{Accuracy}

\[
\text{{Accuracy}} = \frac{{\text{{TP}} + \text{{TN}}}}{{\text{{TP}} + \text{{TN}} + \text{{FP}} + \text{{FN}}}}
\]

\subsection{Recall}

% Recall measures the proportion of actual positive pixels that are correctly identified by the algorithm. The formula for recall in image segmentation is given by:

% Recall formula
\[
\text{{Recall}} = \frac{{\text{{TP}}}}{{\text{{TP}} + \text{{FN}}}}
\]

\subsection{Precision}

% Precision in image segmentation refers to the accuracy of correctly identifying foreground pixels compared to all the pixels classified as foreground. The formula for precision in image segmentation is given by:

% Precision formula
\[
\text{{Precision}} = \frac{{\text{{TP}}}}{{\text{{TP}} + \text{{FP}}}}
\]

\subsection{F1-score}

% The F1-score is a metric commonly used in image segmentation to evaluate the balance between precision and recall. The formula for the F1-score in image segmentation is given by:

% F1-score formula
\[
\text{{F1-score}} = 2 \times \frac{{\text{{precision}} \times \text{{recall}}}}{{\text{{precision}} + \text{{recall}}}}
\]
\vspace{10pt}

\section{Results}
\label{results}

% Gráficos com train x validation loss de cada modelo
% Imagem com comparativo entre as máscaras de diferentes modelos

% The results achieved by each model are displayed on Table \ref{tab:model_metrics}. We can notice that the Mask2Former has outperformed every model in terms of mIoU, Dice, Precision, and F-score. This is most likely associated with its larger number of parameters, which  is also reflected in a mean inference time that is much larger than every one of the other models. 

\begin{figure*}[h]
    \centering
	\includegraphics[width=\textwidth,keepaspectratio]{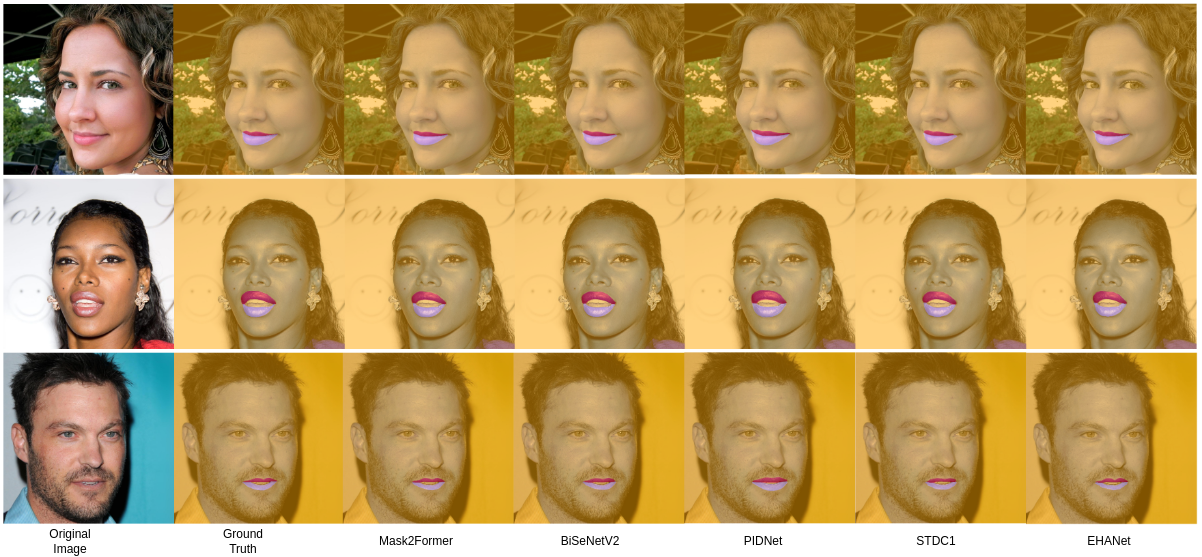}
	\caption{Visualization of masks generated by the models on test samples. Notice that on the first and on the third subject it is much visible that both EHANet and Mask2Former excel at predicting the Cupid's bow region correctly.}
	\label{fig:masks_visualization}
\end{figure*}

Table \ref{tab:model_metrics} presents the mean (and standard deviation) for ten runs of each selected model architecture, as described in the Methodology section.
In the table, each row contains the mean (standard deviation) performance of a model in ten executions of the training process, using different seeds, in multiple metrics.
For instance, EHANet has a mean intersection over union (mIoU) of 82.86\%.
The last column shows the mean inference time, in milliseconds, in a Raspberry Pi4.
The model with the best performance for each metric is highlighted in bold.
To assess the quality of predictions made by each architecture, we have generated masks for selected images from the test set.
These masks are shown in Fig. \ref{fig:masks_visualization}. 

Mask2Former outperforms all other models in terms of absolute mIoU, Dice, and Precision.
This is likely attributed to its larger number of parameters, which is also reflected in a significantly longer mean inference time compared to the other models.
Nonetheless, executing the ANOVA\cite{Fisher1992} test presented a significant difference between all models mIoU's ($p>>1e^{43}$).
Furthermore, by using Tukey's HSD \cite{Tukey1949ComparingIM} test, all pair of models exhibit to be significantly different ($\alpha=0.05$), except for Mask2Former and EHANet ($p\approx0.98$), therefore, it was not possible to find a significant difference between those models.

\begin{table*}[t]
    \caption{Mean (standard deviation) performance of each model using CelebAMask-HQ dataset.}
    \label{tab:model_metrics}
    \centering
    \resizebox{0.9\textwidth}{!}{%
\begin{tabular}{@{}llllllll@{}}
\toprule
\textbf{Model} & \textbf{mIoU} & \textbf{Dice} & \textbf{Recall} & \textbf{Precision} & \textbf{F1-score} & \textbf{Accuracy} & \textbf{\begin{tabular}[c]{@{}l@{}}Mean\\ Inference\\ Time \end{tabular}} \\ \midrule
EHANet~\cite{luo2020ehanet} & 82.86\% (0.25\%) & - & - & - & \textbf{93.72\%} (0.09\%) & 93.94\% (0.27\%) & 1842 (288) \\
Mask2Former~\cite{cheng2022masked} & \textbf{82.96\%} (0.59\%) & \textbf{90.68\%} (0.35\%) & 91.14\% (0.76\%) & \textbf{90.25\%} (0.54\%) & 90.68\% (0.35\%) & 91.10\% (0.78\%) & 15677 (986)\\
BiSeNet V2~\cite{yu2021bisenet} & 80.86\% (0.12\%) & 89.41\% (0.07\%) & 88.83\% (0.38\%) & 89.97\% (0.29\%) & 89.40\% (0.74\%) & 88.85\% (0.40\%) & 3104 (264) \\
PIDNet~\cite{xu2023pidnet} & 72.85\% (0.53\%) & 84.28\% (0.36\%) & \textbf{96.84\%} (0.33\%) & 74.63\% (0.40\%) & 84.28\% (0.36\%) & \textbf{96.84\%} (0.33\%) & \textbf{1781} (108) \\
STDC1~\cite{fan2021rethinking} & 77.07\% (0.36\%) & 87.01\% (0.23\%) & 85.93\% (0.77\%) & 87.93\% (0.77\%) & 87.01\% (0.23\%) & 85.93\% (0.77\%) & 1853 (205) \\ \bottomrule
\end{tabular}
}
\end{table*}

Among the remaining models, which exhibit inference times ranging from 1000 to just over 3000 milliseconds, we observe that EHANet achieves the second-best mean mIoU value, coming close to that achieved by Mask2Former (82.86\% vs. 82.96\%).
For the Dice metric, BiSeNet V2 performs as the second-best model (89.41\% vs. 90.68\% of Mask2Former).
PIDNet achieves the highest mean Recall (96.84\%), but it also exhibits the lowest mean Precision (74.63\%).
The second-best mean Precision is achieved by BiSeNet V2 (89.97\% vs. 90.25\% of Mask2Former).
EHANet attains the highest mean F1-score (93.72\%), and PIDNet achieves the highest mean Accuracy (96.84\%).

Regarding execution time, as mentioned earlier, Mask2Former demonstrates significantly slower performance compared to other architectures. The fastest model is PIDNet, with an average inference time of 1,781 milliseconds on a Raspberry Pi4, on ten executions.

Observing Fig. \ref{fig:masks_visualization} it is evident that the architectures achieving the highest metric scores exhibit better adherence to the contours indicated by the ground truth annotations. This improvement is particularly noticeable in the Cupid's bow region, characterized by its intricate curvature.

Another challenging region to accurately segment is the mouth corners. This difficulty arises due to variations in shading that occur in this region, resulting in pixels with colors different from those typically associated with the lips by the model.

Furthermore, it should be noted that the lip region appears relatively small compared to the background in the images, likely due to the distance at which the images were captured. Consequently, it becomes challenging to discern with clarity whether the segmentation has been accurately delineated or not.

\begin{figure}[!h]
    \centering
	\includegraphics[width=1\columnwidth,keepaspectratio]{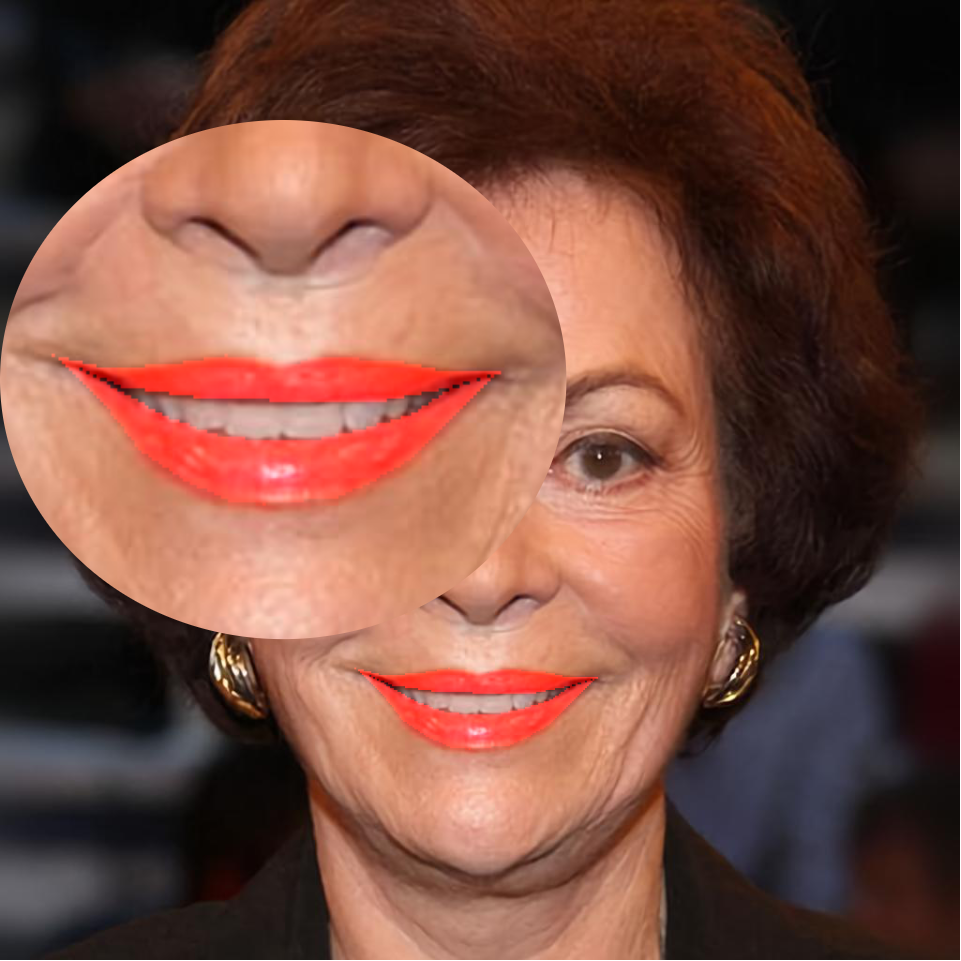}
	\caption{Sample from the CelebAMask-HQ with the label mask colored in red. Notice that from a close look, the segmentation is far from ideal, having a very pixelated aspect and some wrongly labeled pixels. It is also noticeable that a portion of the lower lips is not included in the mask.}
	\label{fig:celebamask_bad_label}
\end{figure}

\vspace{10pt}
\section{Discussion}
\label{discussion}

% Despite achieving better beating nearly all other models on the evaluated metrics, Mask2Former has the drawback of a large prediction time which may result in it being infeasible for certain IoT or edge devices  applications. We can also observe that Mask2Former has displayed certain instability showcased by its high standard deviation on many metrics. In comparison, EHANet has a comparable performance with the advantage of having a relatively small inference time. It also has displayed excellent stability in terms of standard deviation between different evaluations which may indicate that this model has a consistent learning curve and eases the reproduction of results between experiments. This makes EHANet a good fit for low computing power settings where lip mask segmentation is demanded.

Despite achieving interesting absolute values in three metrics, Mask2Former has the drawback of a large prediction time which may result in it being infeasible for certain IoT or edge devices  applications.
We can also observe that Mask2Former has displayed certain instability showcased by its high standard deviation on many metrics. 
In comparison, EHANet did not present a significant mIoU difference and has the advantage of having a smaller inference time. 
It also has displayed better stability in terms of standard deviation between different evaluations which may indicate that this model has a consistent learning curve and eases the reproduction of results between experiments. 
This makes EHANet a good fit for low computing power settings where lip mask segmentation is demanded.

The PIDNet achieved the highest Recall (96.84\%) and Accuracy (96.84\%), but it obtained the lowest Precision (74.63\%). This may indicate that this architecture is able to capture the entire region of the lips in the image, but it extends beyond the lip boundaries. As a result, it has a high number of True Positives and a very low number of False Negatives. However, many pixels that should be classified as background are included in the label space (foreground), causing the denominator of Precision to be large. This is also due to the fact that the number of pixels classified as background is significantly larger than the number of pixels classified as any of the labels, by more than an order of magnitude. This allows predictions that considerably exceed the mask region to appear better than they actually are. One way to mitigate this problem would be to crop the image to a region of interest closer to the lip region. In applications where the user's positioning is relatively fixed, this could be an interesting approach. This measure can help prevent imbalance and under-learning of the model.

During our experiments, we have also noticed that some of CelebAMask-HQ labels display an unreliable lip segmentation mask. This is visible on samples such as the one displayed on Fig. \ref{fig:celebamask_bad_label} where it is noticeable that the segmentation mask is very irregular and pointy. Some random non-contiguous pixels that have a lip label assigned to them are also visible. While these problems may come off as not very relevant when one is considering the dataset as a hole, it may be a drawback when one is trying to get an accurate mask representation for the lips.

\vspace{10pt}

\section{Conclusion}
\label{conclusion}
Towards a lightweight approach to lip segmentation, there is a long path to tread. We have seen that despite many significant progress in the field, substantial challenges remain. Our exploration of EHANet, Mask2former, BiseNet V2, PIDNet, and STDC1 showed that most of these architectures provide robust contour detection under various conditions, including different lighting scenarios, skin tones, lip shapes, and perspectives. However, the computational power required by these models could be a limitation for real-time applications and deployment on IoT devices, where resources are limited. %Therefore, future research efforts should focus on finding novel architectures capable of working with a reduced complexity while retaining the ability to have a robust performance.

As a future direction, we also acknowledge that despite the favorable results, there is still room for further performance improvement, as none of the architectures achieved a performance above 90\% in mIoU. In addition, there is a need for lip-focused datasets designed explicitly for lip segmentation. Although the CelebAMask-HQ dataset is extensive and adequately annotated, there are inconsistencies in the labeling, particularly in the lip area. Some labels may miss parts of the lips or include unrelated areas, potentially hindering performance. Future research efforts should concentrate on discovering novel architectures that can operate with reduced complexity while maintaining robust performance.

\vspace{10pt}

% \section*{Acknowledgment}

% The preferred spelling of the word ``acknowledgment'' in America is without 
% an ``e'' after the ``g''. Avoid the stilted expression ``one of us (R. B. 
% G.) thanks $\ldots$''. Instead, try ``R. B. G. thanks$\ldots$''. Put sponsor 
% acknowledgments in the unnumbered footnote on the first page.

% Please number citations consecutively within brackets \cite{al2020emodet2}. The 
% sentence punctuation follows the bracket \cite{tomczak2014need}. Refer simply to the reference 
% number, as in \cite{cohen2013statistical}---do not use ``Ref. \cite{cohen2013statistical}'' or ``reference \cite{cohen2013statistical}'' except at 
% the beginning of a sentence: ``Reference \cite{cohen2013statistical} was the first $\ldots$''
% Pioneered in a paper by Einstein \cite{einstein}

\bibliographystyle{IEEEtran}
\bibliography{ref.bib}

\end{document}